\pdfoutput=1
\documentclass[letterpaper, 10 pt, conference]{ieeeconf}

\IEEEoverridecommandlockouts
\usepackage{graphicx}
\usepackage{amsmath}
\usepackage{amssymb}
\usepackage{booktabs}
\usepackage{url}

\newcommand{\sysname}{RebarSim}
\newcommand{\Sysname}{RebarSim}

\title{\LARGE \bf
Visual Sim-to-Real Learning for Robotic Insertion under Geometric Variations: Application to Rebar Installation
}

\author{Tao Sun$^{1*}$, Beining Han$^{2*}$, Patrick Yin$^{3}$, Rui Xu$^{1}$,\\
Harry He$^{1}$, Abhishek Gupta$^{3}$, Szymon Rusinkiewicz$^{2}$, Yi Shao$^{1}$\\
$^{1}$McGill University \quad $^{2}$Princeton University \quad $^{3}$University of Washington%
\thanks{This work has been submitted to the IEEE for possible publication. Copyright may be transferred without notice, after which this version may no longer be accessible.}%
\thanks{$^{*}$Equal contribution.}%
\thanks{Corresponding author: Yi Shao (yi.shao2@mcgill.ca).}%
}

\begin{document}

\maketitle
\thispagestyle{empty}
\pagestyle{empty}

\begin{abstract}
Rebar insertion is among the most repetitive and physically
demanding tasks on construction sites, and a contact-rich problem at
1.4\,mm clearance. The parts, however, vary at two levels: a nominal
design per structural member, and fabrication tolerance around each nominal design.
Real-world data therefore has to be re-collected as designs and batches
change. We present
\sysname{}, a visual sim-to-real system trained entirely in simulation. A
privileged state-based teacher is trained with reinforcement learning over
procedurally generated rebar geometries, then distilled into a multi-view
student that maps raw RGB and proprioception directly to actions under
extensive domain randomization. The student transfers to the real world zero-shot,
seating rebars taken from a real factory production run in 91.3\% of real-robot rollouts.
Underlying that result, geometry diversity and pretraining both bring
benefits. Training across a diverse set of nominal designs rather than one lifts
the zero-shot success of both the teacher and the student on unseen designs, and the student policy
outperforms a single-design specialist on that specialist's own design. A pretrained student then
adapts to a new design with 4--6$\times$ fewer distillation samples
than one trained from scratch. Visual sim-to-real transfer depends on appearance randomization and the DAgger
mixture: removing either one sharply lowers success. Videos, code, and task assets are available at
\url{https://rebarsim.github.io}.
\end{abstract}

\section{INTRODUCTION}
\begin{figure}[h]
  \centering
  \includegraphics[width=\columnwidth]{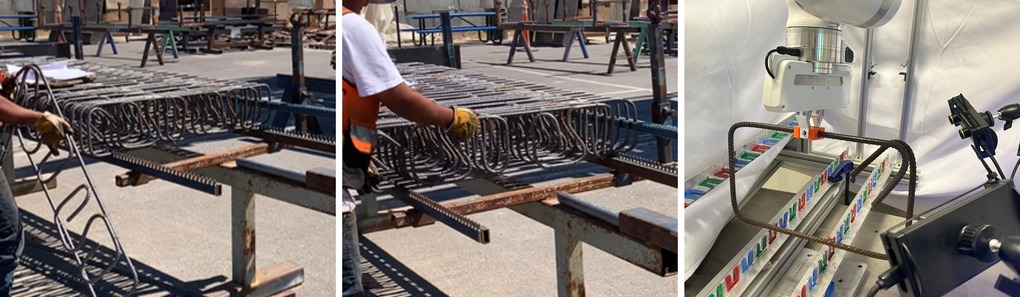}
  \caption{
  Manual rebar insertion on site (left, middle): the worker lifts each rebar,
  then forces it down into the slots, thousands of times a day. Our robot
  setup for the same task (right).}
  \label{fig:motivation}
\end{figure}

On construction sites, rebar-cage assembly remains largely manual (Fig.~\ref{fig:motivation}). Workers repeatedly lift rebars, align them with narrow slots in support racks or other rebars, and insert them into place, often thousands of times per day. This repetitive work is costly, physically demanding, and associated with musculoskeletal injuries~\cite{forde2005ironworkers}, while the construction industry faces a growing shortage of skilled labor~\cite{agc2025workforce}. Automating rebar insertion is therefore an important step toward scalable robotic rebar-cage assembly.

Rebar insertion is also a challenging manipulation problem. Unlike conventional industrial insertion of a fixed, standardized part, a construction site encounters a family of rebar geometries. Different structural members require different nominal designs, while fabrication imprecision further introduces variations in corner angles, cut lengths, out-of-plane twists, and hook geometry (Fig.~\ref{fig:rebar_family}). The rebar's in-hand pose also varies from trial to trial. Across all of this variation, the robot must still achieve millimeter-scale alignment and resolve contact during insertion.

Existing approaches do not directly address this setting. Previous rebar insertion work~\cite{sun2026mobile} learns from real demonstrations, making it costly to scale training across geometric variation. Simulation makes such variation inexpensive through procedural generation and massively parallel experience collection~\cite{mittal2023orbit,narang2022factory}. However, prior sim-to-real insertion methods that handle multiple part geometries~\cite{tang2023industreal,tang2024automate} rely on state-based policies and accurate object pose estimates at deployment. These assumptions are difficult to satisfy for fabricated rebars, whose geometry and in-hand pose are both uncertain. OmniReset~\cite{yin2026omnireset} instead transfers an end-to-end RGB insertion policy from simulation to the real world without pose estimation. However, it is validated only on laboratory peg-in-hole tasks, and its generalization across object geometries is not evaluated.

This work therefore answers whether an end-to-end visual policy that generalizes across both sources of variation can be trained entirely in simulation and transferred to the real world. We introduce \sysname{}, which extends the existing visual sim-to-real paradigm to this setting and learns insertion without real-world demonstrations or an intermediate object-pose estimation stage. To our knowledge, this is the first demonstration of an end-to-end visual sim-to-real policy for object insertion across variations in both object geometry and in-hand pose on a practical industrial task. \Sysname{} achieves 91.3\% success over 150 real-world insertion trials spanning geometric variations and randomized in-hand poses. Simulation experiments further show zero-shot generalization to designs beyond the training distribution and fast adaptation to new designs through fine-tuning.

\Sysname{} follows a teacher-student paradigm~\cite{lee2020learning,chen2021system,handa2023dextreme}. We train a privileged state-based teacher with reinforcement learning over procedurally generated geometries and randomized object initializations, and distill it into a multi-view RGB policy using DAgger~\cite{ross2011reduction}.

Our contributions are:
\begin{enumerate}
\item To our knowledge, we propose the first pipeline that learns an end-to-end visual sim-to-real policy that can perform object insertion across variations in both object geometry and in-hand pose, without real-world demonstrations, validated on a practical rebar insertion task.
\item A study of how training-time geometry diversity affects generalization, showing that it improves robustness within the training range and reduces the adaptation required for geometries beyond it.
\item An empirical study of the visual sim-to-real recipe, showing that visual appearance randomization and the DAgger mixture are key ingredients for successful real-world transfer.
\end{enumerate}

\section{RELATED WORK}
\label{sec:related}

\textbf{Sim-to-Real for Robotics Assembly.} Robotic assembly is a long-standing problem in robot learning~\cite{lee2019making}. Many works learn from real-world teleoperated demonstrations, either by
training a behavior cloning policy from scratch~\cite{zhao2023learning} or by
fine-tuning a generalist vision-language-action (VLA) model~\cite{kim2024openvla}. These methods
require a large number of real demonstrations and often generalize
poorly beyond the demonstrated distribution. Simulation-based RL removes the need for real-world data, and domain
randomization~\cite{tobin2017domain,peng2018sim,akkaya2019solving} and
teacher-student distillation~\cite{lee2020learning,kumar2021rma,handa2023dextreme,
chen2023visual,lum2024dextrah,he2025hover} transfer the resulting policy to the real robot.

Previous works in sim-to-real robot assembly learning differ in what their policies observe.
IndustReal~\cite{tang2023industreal} and AutoMate~\cite{tang2024automate} policies consume 6D object poses as input, but require accurate real-world pose estimation with known CAD models. However, in our problem, fabricated rebars are not standard. In fact, almost no rebar matches its nominal drawing exactly, which weakens the assumption of prior works. Our policy instead reads raw RGB images. OmniReset~\cite{yin2026omnireset} recently transferred RGB visual policies zero-shot on insertion tasks using a broad set of reset states and heavy domain randomization. However, it is limited to simple peg-in-hole tasks, while we address a practical construction task. 

\textbf{Automating Rebar Cage Assembly.} Prior work on automating rebar cage assembly has explored
primitives such as rebar grasping, placing, and tying. Most of the work addresses
\emph{tying}. Typically, the pipeline first recognizes the intersection, then estimates its explicit pose, and plans the tying gun's trajectory, often on gantry- or track-mounted hardware~\cite{jin2021binding,fan2026opentie,momeni2022cages}. These works assume that rebars are already in place on the rack and that their poses can be
recovered by explicit geometric modeling. Our work focuses on the placing primitive.

The work closest to ours~\cite{sun2026mobile}
performs insertion of a tight-fit rebar slot on a mobile manipulator by combining
visual servoing with imitation learning over prompt-segmentation
masks~\cite{ravi2024sam2}. However, it focuses on similar rebar shapes, and the robustness of the policy is not tested against large rebar shape variation. Moreover, it is costly to scale the method to various rebar geometries. \Sysname{} addresses this gap. We learn the same task
from simulation alone, so geometric variation is covered by procedurally generated training rebars.

\section{METHOD}
\label{sec:method}

\begin{figure}[t]
  \centering
  \includegraphics[width=\columnwidth]{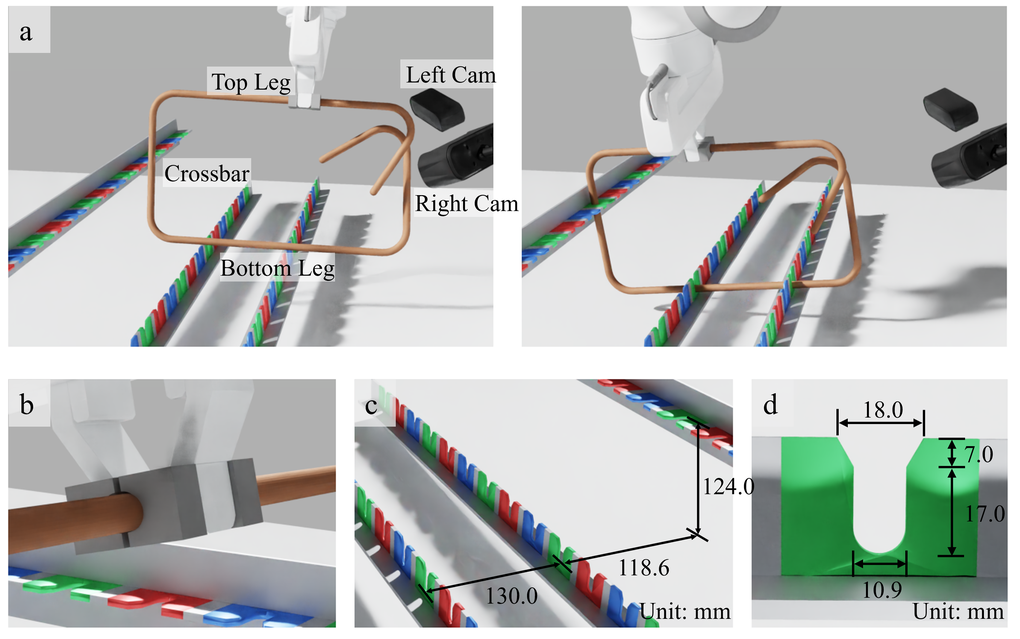}
  \caption{
  Task overview. (a) the robot inserts the rebar it holds into a three-slot
  rack, guided by the left and right cameras and its proprioception. (b) the
  gripper fingers. (c) rack dimensions. (d) slot dimensions.}
  \label{fig:task}
\end{figure}

Our RL environment and training are implemented in Isaac Lab~\cite{mittal2023orbit}.
The robot begins each episode holding a rebar (a steel rod bent into a
closed rectangular loop) and must seat it in three designated slots of a
support rack with 1.4\,mm clearance (Fig.~\ref{fig:task}). The gripper holds one
leg of the loop, which we call the top leg; the opposite leg (the bottom leg)
seats in two low slots and the crossbar, the branch opposite the
hooks, in a third, higher slot (Fig.~\ref{fig:task}). Here, we assume that the rebar is already in hand, with a random gripping pose within a reasonable range. Rack slots are standardized industrial hardware, so the geometry
is fixed and known. Let $\delta_k$ denote the rebar's remaining distance to its seated
pose at slot $k$, measured along the insertion direction. An episode succeeds
when $\max_k \delta_k \le \delta^{\star} = 5$\,mm.

\begin{figure*}[t]
  \centering
  \includegraphics[width=\textwidth]{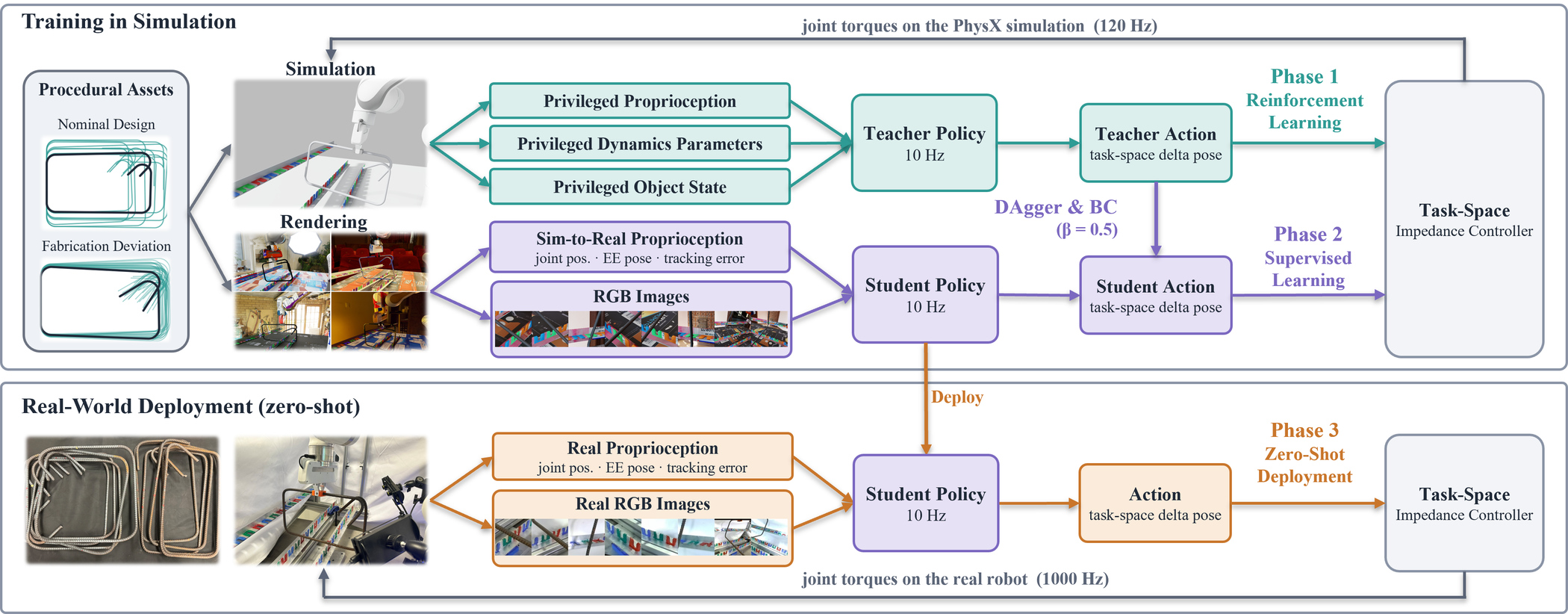}
  \caption{
  Overview of \sysname{}. \emph{Training in simulation:} a procedural
  generator generates the rebar assets. A privileged
  teacher is trained with RL (Phase~1, teal) and supervises a student that
  sees only rendered RGB and proprioception (Phase~2, purple).
  \emph{Deployment:} the same student, frozen, on the real cameras
  (Phase~3, orange).}
  \label{fig:pipeline}
\end{figure*}

\Sysname{} has three development stages (Fig.~\ref{fig:pipeline}): a privileged
\emph{teacher} learning, a visual \emph{student} distillation, and zero-shot
\emph{real-world deployment}. The policy emits a task-space delta pose
$\mathbf{a}_t \in \mathbb{R}^{6}$ of the fingertip frame at 10\,Hz. Joint
torques are computed with a task impedance controller
(Sec.~\ref{sec:method:teacher}). The teacher
$\pi_{\text{teacher}}(\mathbf{a}_t \mid o_t^{\text{priv}})$ acts on privileged
simulator state, and the student $\pi_{\text{student}}(\mathbf{a}_t \mid
o_t^{\text{vis}})$ reads $o_t^{\text{vis}} = (\mathbf{I}_t^{1:K},
o_t^{\text{prop}})$: $K{=}8$ calibrated RGB views and the proprioception
$o_t^{\text{prop}}$ defined in Sec.~\ref{sec:method:student}.

\subsection{Simulation with Geometric Variation}
\label{sec:method:sim}

We cast the task as a finite-horizon POMDP whose dynamics are parameterized
by a task instance
\begin{equation}
\xi = \big(g,\; \mathbf{s}_0\big) \sim p(\xi),
\qquad g = (d, \boldsymbol{\epsilon}) \in \mathcal{G},
\end{equation}
resampled at every reset in simulation. Here $g$ is the rebar geometry, a
nominal design $d$ with fabrication deviation $\boldsymbol{\epsilon}$, and
$\mathbf{s}_0$ is the initial scene state. This subsection describes how we
generate $\mathcal{G}$, and Sec.~\ref{sec:method:resets} how we generate
$\mathbf{s}_0$.

\begin{figure}[t]
  \centering
  \includegraphics[width=\columnwidth]{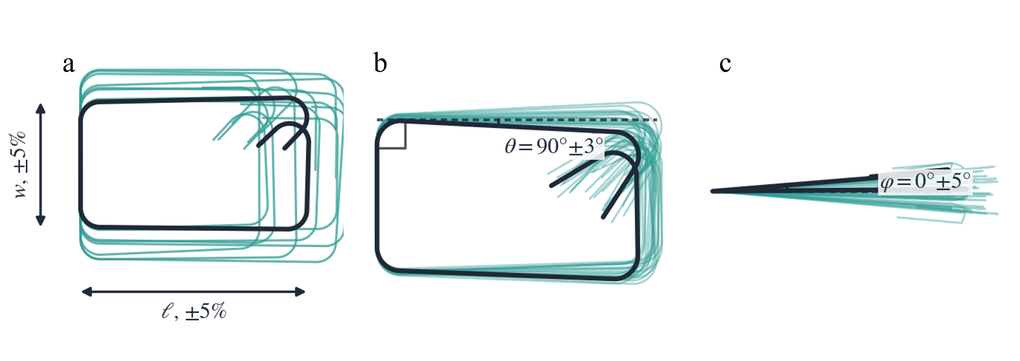}
  \caption{
  The rebar generator's parameters. (a) the
  lengths $\ell$ and $w$. (b) the corner angle $\theta$. (c) the
  out-of-plane twist $\varphi$.}
  \label{fig:rebar_family}
\end{figure}

We scope $\mathcal{G}$ to \emph{stirrups} (the transverse rebars wrapping the
longitudinal steel of essentially every concrete beam, column and pile cage,
and among the most numerous rebar shapes on site) in the closed rectangular
form of Fig.~\ref{fig:rebar_family}.

\textbf{Rebar Generator.} We design our rebar generator to cover a wide range of rebar geometries in training.
A nominal design is $d = (\ell, w)$, the leg length
and the width of the crossbar. Fabrication then perturbs it by
$\boldsymbol{\epsilon} = (\epsilon_\ell, \epsilon_w,
\{\theta_i\}_{i=1}^{3}, \{\varphi_i\}_{i=1}^{3}, h, \alpha)$: the leg length
$\ell$ and the width $w$ each at $\pm5\%$, each of the three straight branches (two legs and the crossbar) at
corner angle $\theta_i = 90^\circ\!\pm\!3^\circ$ and out-of-plane twist
$\varphi_i = 0^\circ\!\pm\!5^\circ$, and the two hooks sharing a length $h \in
[20, 100]$\,mm and an angle $\alpha \in \{45^\circ, 60^\circ, 90^\circ\}$.
The parameters and their ranges come from standard fabrication practice
and from the rebars we measured. Our experiments use \#3 rebar, a common size
on site, with the diameter held at its 9.5\,mm nominal, since rolling
tolerance is far tighter than cutting and bending tolerance.

\textbf{Physics Modeling.} Both rebars and the rack use signed-distance-field (SDF)
collision detection. This work studies the insertion skill itself, with the
target fixed to the center slot group, so only that group carries SDF
collision geometry. Selecting the target group, for example from a color
or position cue in the observation, is left to future work. We simulate the in-hand grasp as a revolute hinge
along the hand's x-axis with friction. This simplification reduces simulation instability from thin rebars but still allows sim-to-real transfer.

\subsection{Reset-State Bank Generation}
\label{sec:method:resets}

Reset states covering every stage of a task let on-policy RL
master long-horizon, contact-rich skills with simple
rewards~\cite{yin2026omnireset}. Ours randomize the outcome of a preceding
grasp: where the rebar sits in the hand, and where the hand sits
relative to the rack. We precompute two families of reset states, mixed
1:1 during training, at 5{,}000 states per rebar variant.

\textbf{Free-space approach states} place the grasped rebar uniformly in a box
anchored at the variant's success pose and expressed in the rack frame
(up to 5\,cm horizontally withdrawn from the seated pose, $\pm$3\,cm lateral, up to 5\,cm above), with
per-axis orientation jitter of $\pm$0.17\,rad.

\textbf{Partially inserted states} are generated by \emph{physics back-out}: with gravity off, randomized wrenches push the rebar
from the success pose along its low-resistance directions, and every
still-engaged intermediate pose is recorded. We reject-sample
to $\approx$1/3 each for 3, 2 and 1 engaged slots, and drop states that are
already seated.

Each recorded state additionally carries a randomized arm posture, grasp and
rack placement. IK over this confined workspace returns configurations from a
narrow band, so each solution is perturbed inside the nullspace of the task
Jacobian, moving the elbow but not the rebar. The in-hand roll
($\pm$0.12\,rad), the grip slide along the top leg's axis ($\pm$3\,cm) and the rack
pose ($\pm$3\,cm, $\pm$0.15\,rad) are sampled once at generation time, at ranges that
cover the range an upstream picking stage such as~\cite{sun2026mobile} produces.

\subsection{Privileged Teacher Training}
\label{sec:method:teacher}

\textbf{Teacher formulation.} We formulate the teacher as a state-based RL
policy $\pi_{\text{teacher}}(\mathbf{a}_t \mid o_t^{\text{priv}})$ trained
with PPO~\cite{schulman2017proximal}. Actor and critic are separate MLPs of
width $[512, 256, 128, 64]$. The privileged observation stacks the previous
action, proprioception, the rebar's roll in the grip, and the rebar and rack poses
in the rack frame over the last five control steps, plus two task-specific
terms. $\mathbf{x}_t^{\text{seg}}$, the pose of the bottom leg and of the
crossbar (Fig.~\ref{fig:task}) in the rack frame, exposes the sampled geometry as segment poses (midpoint and direction)
rather than as shape parameters. $\mathbf{f}_t$, the peak rebar--rack contact
force and its excess over the penalty threshold, lets the policy modulate
contact rather than only being penalized for it.

\textbf{Action Space.} The action $\mathbf{a}_t = (\Delta\mathbf{x}_t,
\Delta\boldsymbol{\omega}_t) \in \mathbb{R}^{6}$ is a Cartesian pose delta
relative to the \emph{current} fingertip pose, scaled by 0.04\,m and
0.06\,rad per step and tracked over the seven arm joints by a task
impedance controller,
\begin{equation}
\boldsymbol{\tau} = \mathrm{clamp}\!\left(J^{\top}\!\left[K_p\,
\mathbf{e}_t - K_d\,\mathbf{v}\right]\right),
\end{equation}
where $\mathbf{e}_t$ is the fingertip pose tracking error. Note that the clamp acts on the joint torque, not on $\mathbf{a}_t$. The critic additionally
observes privileged quantities: remaining episode time, velocities, and the
dynamics parameters sampled by the randomization below.

\textbf{Reward Design.} The reward is a fixed weighted sum $r_t = \sum_i \lambda_i
r_i$ of the terms in Table~\ref{tab:reward}, shared unchanged by every
geometric variant, with no staged shaping. The only schedule applies to the
contact term, whose magnitude is ramped up linearly over the first 300
iterations.

Here $\Delta p_t$ and $\Delta\psi_t$ are the position and orientation deviation of the rebar
from its per-variant seated pose, computed analytically in the rack frame,
with scales $\sigma_p{=}0.10$\,m and $\sigma_a{=}0.60$\,rad. The contact
term penalizes only the rebar--rack force $F_{t,j}$ above $F_0{=}30$\,N, averaged
over the $T{=}12$ physics substeps $j$ per control step and normalized by
$F_{\text{ref}}{=}100$\,N, so that pressing hard is discouraged.

\begin{table}[t]
\caption{Reward terms and weights.}
\label{tab:reward}
\begin{center}
\begin{tabular}{llr}
\toprule
Term & Expression & $\lambda_i$ \\
\midrule
Alignment      & $\tfrac{1}{2}\!\left[e^{-\Delta p_t/\sigma_p} + e^{-\Delta\psi_t/\sigma_a}\right]$ & $0.1$ \\
Success        & $\mathbf{1}\!\left[\max_k \delta_k \le \delta^{\star}\right]$        & $1.0$ \\
Contact force  & $\tfrac{1}{T F_{\text{ref}}}\sum_{j} \mathrm{clamp}\!\left(\lVert F_{t,j} \rVert - F_0, 0, 200\,\text{N}\right)$ & $-0.05$ \\
Action mag.    & $\lVert \mathbf{a}_t \rVert^2$                                        & $-10^{-4}$ \\
Action rate    & $\lVert \mathbf{a}_t - \mathbf{a}_{t-1} \rVert^2$                     & $-10^{-3}$ \\
Arm joint vel. & $\lVert \dot{\mathbf{q}}_t \rVert^2$                                  & $-10^{-2}$ \\
Abnormal arm   & $\mathbf{1}\!\left[\,\lvert \dot{q} \rvert > 2\,\dot{q}_{\max}\right]$ & $-100$ \\
\bottomrule
\end{tabular}
\end{center}
\end{table}

\textbf{Domain Randomization.} Dynamics randomization is anchored at
system-identified nominal values~\cite{peng2018sim,handa2023dextreme}: arm
friction, armature, and task-impedance gains are jittered by $\pm$20\%
around the SysID point, which is provided with the released controller. Surface
friction is drawn per environment: static and dynamic $\mu$ from $U(1.0, 2.0)$ and
$U(0.9, 1.9)$ for both the rebar and the rack, and from $U(0.3, 1.2)$ and
$U(0.2, 1.0)$ for the gripper pads.

\subsection{Visual Student Distillation}
\label{sec:method:student}

The student $\pi_{\text{student}}$ observes proprioception and eight RGB
views: six close-up views centered on the slots, plus two wide views that
keep the whole rebar in frame. Sec.~\ref{sec:method:real} describes how
these views are obtained in simulation and on the real robot.

\textbf{DAgger--BC Distillation.} We distill the privileged teacher into the
student with a hybrid of online DAgger~\cite{ross2011reduction} and behavior
cloning (BC)~\cite{he2025viral}. Both share the same regression objective,
computed over a mixture of the observation distributions induced by teacher
and student rollouts:
\begin{equation}
\mathcal{L}_{\text{distill}} = \mathbb{E}_{o_t \sim \rho^{o}}
  \Big[ \big\| \pi_{\text{teacher}}\big(o_t^{\text{priv}}\big)
  - \pi_{\text{student}}\big(o_t^{\text{vis}}\big) \big\|_2^2 \Big],
\end{equation}
where $\rho^{o} = \beta\,\rho^{o}_{\pi_{\text{teacher}}} +
(1-\beta)\,\rho^{o}_{\pi_{\text{student}}}$ mixes the two rollout
distributions with ratio $\beta$, and $o_t^{\text{priv}}$,
$o_t^{\text{vis}}$ are the privileged and visual observations of the same
simulator state. We use $\beta = 0.5$, following the ratio ablation
of~\cite{he2025viral}.

\textbf{Network Architecture.} One ImageNet-pretrained ResNet-18 encodes all
eight views, and a learnable per-view embedding marks which view a feature
came from. The features are concatenated with the student's proprioception
$o_t^{\text{prop}} = [\,\mathbf{q}_t,\, \mathbf{x}_t^{\text{ee}},\,
\mathbf{e}_t\,]$ (the nine joint positions of the arm and gripper fingers, the fingertip pose in the robot
base frame, and the fingertip tracking error, 21 dimensions carrying no
previous action and nothing about the rebar or the rack) over a four-step
history, and mapped to the action by an MLP with two hidden layers of 512
ReLU units and a linear output layer.

\begin{figure*}[t]
  \centering
  \includegraphics[width=\linewidth]{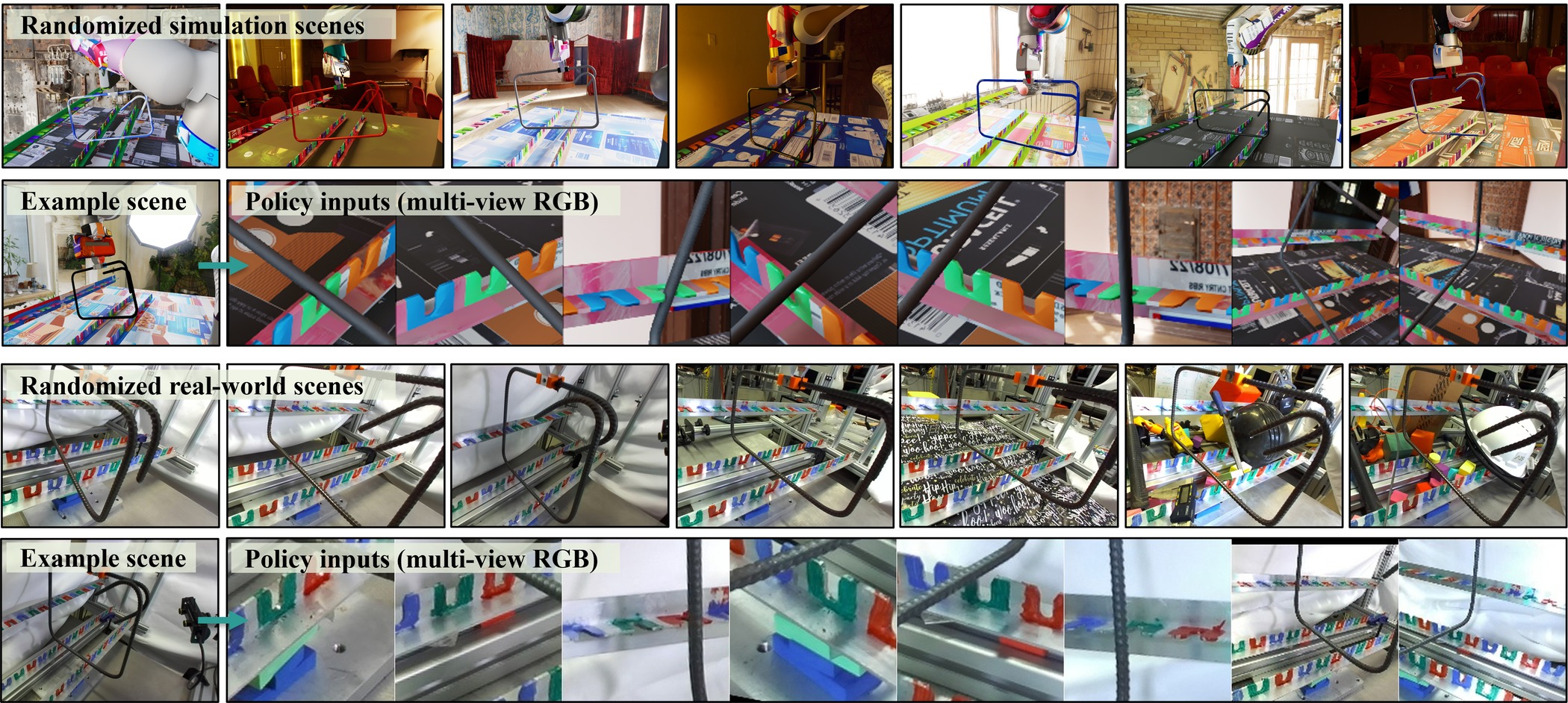}
  \caption{\textbf{Training scenes in simulation (top) and evaluation scenes
  on the real robot (bottom).} The student never sees a real image while it
  is trained: the top block is the distribution it is distilled on, the
  bottom block is the real-world setup it is deployed in. Each block pairs the spread of
  randomized scenes with one example scene and the multi-view RGB the policy
  actually reads in it.}
  \label{fig:scenes} 
\end{figure*}

\textbf{Domain Randomization.} Because the student reads pixels
directly, distillation randomizes everything that changes how the scene
looks. Dome lighting is resampled from a pool of HDRI environment maps. The
rebar, the rack, the table and the robot each receive an independently
sampled texture and surface material (color, roughness and metallic). Camera poses are jittered around the calibrated real setup: $\pm$1\,cm on
the optical center of all eight views, and a further $\pm$3$^\circ$ of
orientation on the two wide ones. Intrinsics are held at their calibrated
values. Geometry and dynamics randomization continue from teacher
training. Fig.~\ref{fig:scenes} (top) shows the spread of scenes this
produces.

\subsection{Real-World Deployment}
\label{sec:method:real}

\textbf{Control.} We use a Franka Research 3 robot. Policy actions are
executed by a task impedance controller of Sec.~\ref{sec:method:teacher},
whose torque is interpolated to 1\,kHz in the real world. In the
real world, the rebars used in our experiments weigh about 0.7\,kg;
we compensate this gravity load on the end effector manually, and leave
automatic compensation to future work.

\textbf{Perception.} Fig.~\ref{fig:pipeline} (bottom) shows our robot setup: two ZED
Mini stereo cameras are mounted to see the rack and the robot's workspace. We recover
their poses from a calibration marker and place the simulated cameras correspondingly.

In our simulation the eight views are rendered by eight cameras. In the real
world, each of the two rectified left images is cut into fixed bounding-box
crops, three close-up views centered on the slots plus one wide view
(Fig.~\ref{fig:scenes}, bottom), and each crop is warped with a homography
into a virtual camera that faces its target as the corresponding simulated
camera does.

\section{EXPERIMENTS}
\label{sec:exp}

We design our experiments to answer the following questions:
\begin{enumerate}
\item[\textbf{Q1}] Does \sysname{} generalize across rebar geometry, in
simulation and on the real robot, with no real-world data?
\item[\textbf{Q2}] How does geometry randomization during training affect
final performance, and how does it affect the speed of adapting to novel
geometries?
\item[\textbf{Q3}] Which design choices of the visual sim-to-real recipe are
critical in practice?
\end{enumerate}

\textbf{Simulation Training.} All policies are trained on four NVIDIA L40
GPUs. The teacher
trains state-only in 16{,}384 environments per GPU. The
student renders its eight $150\times150$ camera views with tiled rendering at 74
environments per GPU. We evaluate on three sets of nominal designs.
$\mathcal{D}_{\text{train}}$ is the $3\times3$ grid $\ell \in \{290, 350,
400\}$\,mm $\times\, w \in \{195, 240, 290\}$\,mm.
$\mathcal{D}_{\text{interp}}$ is four unseen designs inside that grid,
$\ell \in \{320, 375\}$\,mm $\times\, w \in \{217.5, 265\}$\,mm.
$\mathcal{D}_{\text{extrap}}$ is the four corners of $\ell \in \{260, 430\}$\,mm
$\times\, w \in \{150, 320\}$\,mm, which lie outside the grid on both axes. Each
design carries 20 variants, giving 180, 80 and 80, with
$\boldsymbol{\epsilon}$ drawn throughout from the tolerance ranges of
Sec.~\ref{sec:method:sim}, so the design $d$ is the only factor that differs between
the sets.
Every simulation success rate we report is measured with the
deterministic policy over 4096 episodes for the teacher and 2304 for the
student, with rebar variants drawn uniformly at random from the evaluated set. Simulated and real evaluations both
start from a held-out evaluation bank of free-space states
(Sec.~\ref{sec:method:resets}).

\textbf{Real-world Evaluation.} Real rollouts cover two designs, $\mathcal{D}_1$ ($\ell = 350$,
$w = 195$\,mm) and $\mathcal{D}_2$ ($\ell = 290$, $w = 290$\,mm). Both are
grid points of $\mathcal{D}_{\text{train}}$, and for each we use a batch of
rebars from a real factory production run. The rebars deviate markedly from their nominal
geometry: corner angles reach their $3^\circ$ tolerance limit and
out-of-plane twists their $5^\circ$ limit
(Sec.~\ref{sec:method:sim}). Both are scored in simulation \emph{and} on the real robot, so
their gap is a sim-to-real drop of one policy on one design.

\textbf{Task Metric.} We report the success rate, under the $\max_k \delta_k \le
\delta^{\star}$ criterion of Sec.~\ref{sec:method} applied
identically in our simulated and real-robot evaluations.

\subsection{Generalization across Geometric Variations (Q1)}
\label{sec:exp:main}

Table~\ref{tab:main} reports \sysname{} in simulation and on the real robot.
In simulation, it scores 96.1\% on the nine designs it trained on and
96.6\% on $\mathcal{D}_{\text{interp}}$, the four unseen designs held out
inside that grid. That performance transfers to the real robot, where it
seats the rebar in 137 of 150 rollouts (91.3\%): 68/75 on $\mathcal{D}_1$ and
69/75 on $\mathcal{D}_2$, against 95.4\% and 96.8\% for the same policy in
simulation.
We observed no pronounced difference in success rate between
individual rebars. The dominant failure mode is discussed in
Sec.~\ref{sec:exp:qual}, with a more detailed account on the project website.

\begin{table}[t]
\caption{\Sysname{}'s simulated and real rollout results on different rebar
groups. On the real robot, $\mathcal{D}_1$ and $\mathcal{D}_2$ each
contribute five rebars drawn at random from their group, fifteen rollouts per
rebar, with the rebar's initial pose, the grasp pose and the rack pose
randomized as in simulation. Real entries are the success rate (\%), with
successes/rollouts in parentheses.}
\label{tab:main}
\begin{center}
\begin{tabular}{lcc}
\toprule
Evaluation set & Sim SR & Real \\
\midrule
$\mathcal{D}_{\text{train}}$ \ (9 designs)       & 96.1 & --~(sim only) \\
$\mathcal{D}_{\text{interp}}$ (4 unseen designs) & 96.6 & --~(sim only) \\
$\mathcal{D}_{\text{extrap}}$ (4 designs outside grid) & 77.6 & --~(sim only) \\
\cmidrule(lr){1-3}
$\mathcal{D}_1$                          & 95.4 & 90.7 (68/75) \\
$\mathcal{D}_2$                          & 96.8 & 92.0 (69/75) \\
\cmidrule(lr){1-3}
all real rollouts                        & --   & 91.3 (137/150) \\
\bottomrule
\end{tabular}
\end{center}
\end{table}

\begin{figure}[t]
  \centering
  \includegraphics[width=\columnwidth]{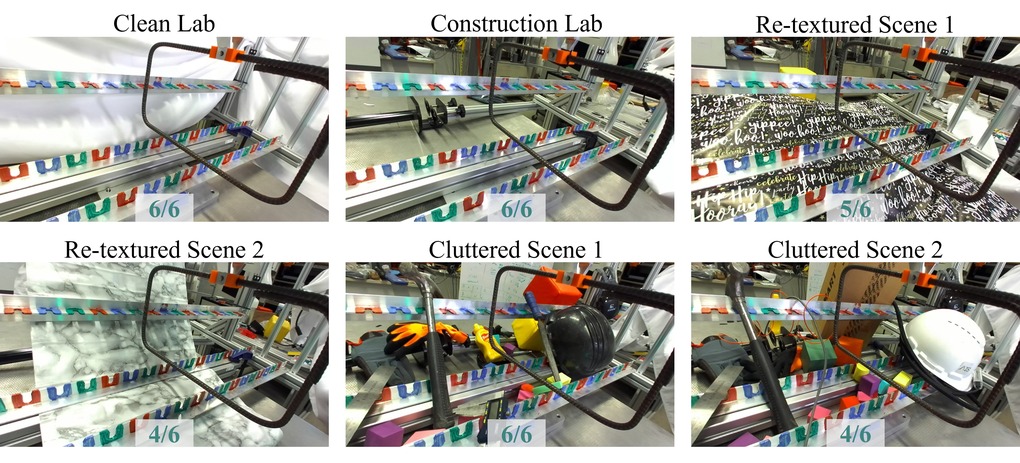}
  \caption{
  Real-robot success across backgrounds. The six rollouts are the same six
  everywhere (two rebars drawn at
  random from $\mathcal{D}_1$, three initial EE poses and rebar-in-hand poses each, held
  fixed as the scene changes), so the panels differ in background alone. The
  five deployed scenes pool to 25/30 against 6/6
  in the clean lab, top left.}
  \label{fig:background}
\end{figure}

\textbf{Background robustness.} Fig.~\ref{fig:background} examines how the
visual background affects the policy on the real robot. Across 30 rollouts in
five deployed scenes (a construction lab, two re-textured and two
cluttered), it seats 25 rebars, against 6 of 6 in the clean lab. Success
degrades rather than collapses: the two hardest scenes, a marble backdrop and
a heavily cluttered scene with toys, tools and cables, still seat 4 of 6.

\subsection{Effect of Geometry Randomization (Q2)}
\label{sec:exp:geomrand}

\begin{figure}[t]
  \centering
  \includegraphics[width=\columnwidth]{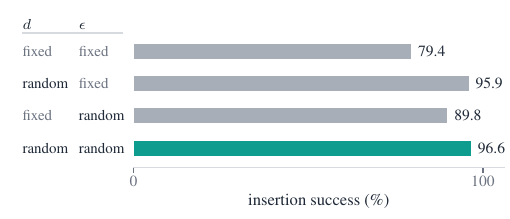}
  \caption{
  Student evaluation on $\mathcal{D}_1$ under the four training cases. For
  each training case, we report the success rate of its distilled student's
  best checkpoint on $\mathcal{D}_1$.}
  \label{fig:geomrand}
\end{figure}

Rebar geometry factors as $g = (d, \boldsymbol{\epsilon})$
(Sec.~\ref{sec:method}), the nominal design and the fabrication
deviation. We build four training cases as a $2\times2$ factorial over which
component of $g$ is randomized. All four train on a set that contains
$\mathcal{D}_1$'s nominal design and all are scored on $\mathcal{D}_1$:
neither (that design as a single exact rebar), $\boldsymbol{\epsilon}$ only
(that design across its tolerances, i.e.\ $\mathcal{D}_1$ itself), $d$ only (the nine exact designs of the
training grid, which include it), and both (those nine across their
tolerances, the set \sysname{} trains on). We train a teacher and its
distilled student for each, and report the student (Fig.~\ref{fig:geomrand}).

Fig.~\ref{fig:geomrand} separates the two main effects. With
$\boldsymbol{\epsilon}$ held fixed, randomizing the design $d$ lifts success
from 79.4\% to 95.9\%. With $d$ held fixed, randomizing the fabrication
deviation $\boldsymbol{\epsilon}$ lifts it to 89.8\%. Both matter, and
randomizing both is best, at 96.6\%. Design diversity has the larger main
effect (16.5 points against 10.4) and the larger marginal effect: adding $d$ on
top of $\boldsymbol{\epsilon}$ gains a further 6.8 points, adding
$\boldsymbol{\epsilon}$ on top of $d$ only 0.7. Diversity along $d$ therefore transfers to
$\boldsymbol{\epsilon}$ while the reverse does not: a policy trained across
the nine exact designs already handles fabrication deviations it
was never trained on, whereas one trained on a single design across its full
tolerance range still trails the nine-design policy by 6.8 points on the very
design it trained on. Design-level diversity also does
not reduce performance on any single design: the nine-design policy is the best of the
four on this one design, a conclusion similar to the object-generalization result
reported for grasping in~\cite{he2025viral}.

\subsection{Adaptation Speed to Novel Geometries (Q2)}
\label{sec:exp:adapt}

We fine-tune on $\mathcal{D}_{\text{extrap}}$ at both
levels of the pipeline. \textbf{Teacher level:} three initializations --- \sysname{}
(teacher randomizing both $d$ and $\boldsymbol{\epsilon}$), the $d$-fixed and
$\boldsymbol{\epsilon}$-random teacher of Sec.~\ref{sec:exp:geomrand}, and
random weights, i.e.\ from scratch --- are fine-tuned on the new designs
and their tolerance ranges under one common budget of 1.0\,B samples, three
seeds each, and compared by the number of samples needed to reach 90\% success. \textbf{Student level:} the
adapted \sysname{} teacher is then distilled into students initialized from
the pretrained \sysname{} student, from the pretrained $d$-fixed student, and
from scratch, three seeds each. All three
distill from that one teacher, so the student's starting weights are the only
difference.

\begin{figure}[t]
  \centering
  \includegraphics[width=\columnwidth]{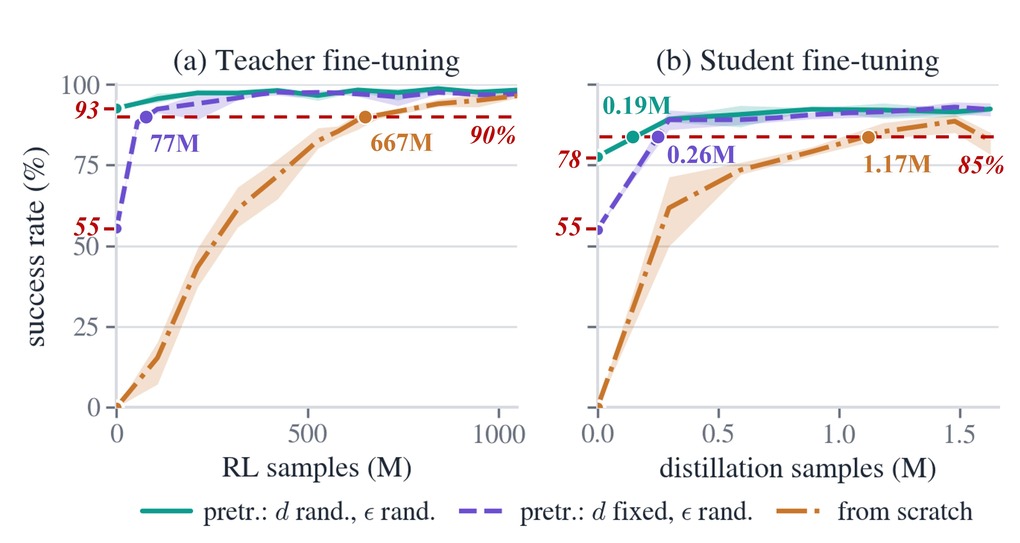}
  \caption{
  Adaptation to $\mathcal{D}_{\text{extrap}}$, the four designs outside the
  training grid. Labels indicate which components of the rebar geometry $g = (d, \boldsymbol{\epsilon})$ the pretraining
  randomized. The $x$-axis counts only adaptation samples, so each curve
  starts at its zero-shot success, marked on the left axis. The line is the
  mean over three seeds, the band their min--max.
  (a) teacher RL fine-tuning, dots at samples-to-90\%.
  (b) student distillation, dots at samples-to-85\%.}
  \label{fig:adapt}
\end{figure}

At the teacher level, shown in Fig.~\ref{fig:adapt} (a), training across
diverse nominal designs raises zero-shot success on the four unseen designs
from 55.4\% for single-design pretraining to 92.6\%, against 0.0\% from
scratch, so the three initializations rank by how many designs they saw in
pretraining. Diversity also shortens adaptation: \sysname{} starts above the
90\% mark (fine-tuning adds a further 5.6 points), whereas the $d$-fixed
teacher needs 77\,M samples to reach it and training from scratch 667\,M.
All three teachers converge to 96--98\% within the shared budget. At the
teacher level, diversity therefore determines how many samples a new design
takes, not the final performance on it.

At the student level, shown in Fig.~\ref{fig:adapt} (b), the dominant factor is whether
the student was pretrained at all. Zero-shot, the \sysname{} student already
reaches 77.6\% on $\mathcal{D}_{\text{extrap}}$, against 55.1\% for the
$d$-fixed student. When adaptation is needed, the \sysname{} and $d$-fixed
students reach 85\% after 0.19\,M and 0.26\,M distillation samples, 6.2$\times$ and 4.5$\times$
fewer than the 1.17\,M a student trained from scratch needs, and they
plateau at 90--93\% over their last four evaluations against the from-scratch
student's 81--85\%. Pretraining therefore both reduces the distillation
samples a new design needs and raises the final success rate.

\subsection{Visual Sim-to-Real Recipe Ablations (Q3)}
\label{sec:exp:recipe}

We ablate three ingredients of the recipe, removing or replacing each one in
turn: appearance randomization, viewpoint randomization and the DAgger
mixture. Table~\ref{tab:recipe} reports the results in simulation on
$\mathcal{D}_{\text{train}}$.

Appearance randomization is indispensable.
The student trained without it still solves its own fixed-appearance
environment (91.1), but collapses to 2.8 under the randomized appearances the
deployed policy is scored on. The other two matter less.
Viewpoint randomization is worth 5.9 points. Replacing the
DAgger mixture with pure BC costs 27.2 points, a drop we attribute to the covariate shift of a student
that is never asked to act from its own states. The mixture keeps the
teacher-to-student drop small on this task.

\begin{table}[t]
\caption{
  Training-recipe ablations, evaluated in simulation on the nine designs of
  $\mathcal{D}_{\text{train}}$. All rows ablate distillation with the
  teacher, reset banks and assets held fixed.}
\label{tab:recipe}
\begin{center}
\begin{tabular}{lc}
\toprule
Recipe                            & Sim SR (\%) $\uparrow$ \\
\midrule
\Sysname{} (full recipe)           & \textbf{96.1} \\
\midrule
w/o appearance randomization       & 2.8 \\
w/o viewpoint randomization        & 90.2 \\
pure BC (no DAgger)                & 68.9 \\
\bottomrule
\end{tabular}
\end{center}
\end{table}

\subsection{Observed Behaviors and Failure Modes}
\label{sec:exp:qual}

\textbf{Emergent recovery.} For rebar insertion, the most common failure is
the rebar colliding with the rack near the slots and getting stuck there.
Such contact states are many and varied, so \cite{sun2026mobile} collects
dedicated failure-recovery demonstrations to handle them, which takes
substantial time and effort. In our real rollouts, about 20\% of the
insertions did not seat the rebar on the first attempt: the rebar first collided with the
rack around the slots, and the policy backed it out and re-attempted the
insertion until the rebar seated (video on the project site). This
back-out-and-retry behavior is not programmed anywhere in the policy or the
controller. It emerges from RL training with a reward that has no retry term
(Table~\ref{tab:reward}).

\textbf{Failed recovery.} In some rollouts the policy does not back the rebar
out after the collision but keeps rotating it to slide it into the slots. The
grasp cannot transmit enough torque to overcome the contact friction, so the
continued rotation produces a growing in-hand roll shift until the rebar can
no longer be recovered. We attribute this to contact differences between
simulation and the real world, which let the policy learn a behavior that is
harder to execute on the real robot.

\section{DISCUSSION AND CONCLUSIONS}
\label{sec:conclusion}

We built and deployed \sysname{}, a visual sim-to-real system for
contact-rich insertion under geometric variation, instantiated on rebar
insertion. Trained entirely in simulation over procedurally generated
geometric variants and deployed directly from raw camera images, it seats
rebars taken from a real factory production run at a 91.3\% success rate over
150 real-robot rollouts, with no real-world data anywhere in the pipeline. Our
ablations trace that result to a few design choices. Geometry diversity and
pretraining both bring benefits: training across nine nominal designs rather
than one raises success on the training designs and lifts zero-shot success
outside them from 55.4\% to 92.6\%, while pretraining of either kind
makes a new design sample-efficient to acquire, at 4--6$\times$ fewer distillation
samples than starting from scratch. The visual recipe rests on appearance
randomization, without which the student does not transfer at all, and the
DAgger mixture, worth 27.2 points over pure behavior cloning.

The deployed system has two limitations, and both point to replacing a
hand-designed component with one learned in simulation. First,
\sysname{} assumes the rebar is already grasped, and acquiring that grasp is
left to a preceding module or policy, so we demonstrate only the
insertion stage. We intend
to train that stage in simulation with RL as well, and compose the two into
a single grasp-and-insert pipeline. Second, the student's viewpoints are fixed
at design time: its eight views are fixed bounding-box crops
(Sec.~\ref{sec:method:real}) laid out for the target slot group, and other
slot groups on the rack are not considered. Mounting the robot on a mobile
base could keep the cameras aligned with whichever slot group is the target,
but this limits how conveniently the system can be deployed. An active-vision
mechanism, letting the policy choose which region of the image to resolve,
would make that choice part of the learned policy.

\end{document}